\documentclass[papersize,journal]{IEEEtran}
\usepackage{amsmath,amsfonts}
\usepackage[ruled,vlined]{algorithm2e}
\usepackage{array}
\usepackage{booktabs}
\usepackage{caption}
\usepackage{graphicx}
\usepackage{siunitx}
\usepackage{stfloats}
\usepackage{url}
\usepackage{verbatim}
\usepackage{xcolor}
\usepackage{listings}
\usepackage{balance}
\usepackage{colortbl}
\usepackage{bm}
\usepackage{amssymb}

\usepackage{tabularx} 

\usepackage[caption=false,font=normalsize,labelfont=sf,textfont=sf]{subfig}
\usepackage{hyperref}
\hypersetup{
  colorlinks=true,
  linkcolor=black,
  filecolor=black,
  urlcolor=blue,
  citecolor=blue,
  pdftitle={Paper Omitted for Review}
}

\def\BibTeX{{\rm B\kern-.05em{\sc i\kern-.025em b}\kern-.08em
    T\kern-.1667em\lower.7ex\hbox{E}\kern-.125emX}}

\begin{document}

\title{RAPTAR: Radar Radiation Pattern Acquisition through Automated Collaborative Robotics}

\author{
Maaz Qureshi$^{1,*}$, Mohammad Omid Bagheri$^{2}$, Abdelrahman Elbadrawy$^{2}$, William Melek$^{1}$, and George Shaker$^{2}$



\thanks{$^{1}$Maaz Qureshi and William Melek are with the Department of Mechanical and Mechatronics Engineering, University of Waterloo, ON N2L 3GL, Ontario, Canada. $^{*}$Co-responding Author E-mail: m23qures@uwaterloo.ca.}%
\thanks{$^{2}$Mohammad Omid Bagheri, Abdelrahman Elbadrawy, and George Shaker are with the Department of Electrical and Computer Engineering, University of Waterloo, ON N2L 3GL, Ontario, Canada.}%
\thanks{GitHub: \url{https://github.com/Maaz-qureshi98/RAPTAR-Motion-Planning.git}}%

}

\markboth{IEEE Robotics and Automation Letters (RA-L) Submission}
{Paper Omitted for double-blind peer review policy}

\maketitle

\begin{abstract}

Accurate characterization of modern on-chip antennas remains challenging, as current probe-station techniques offer limited angular coverage, rely on bespoke hardware, and require frequent manual alignment. This research introduces RAPTAR (Radiation Pattern Acquisition through Robotic Automation), a portable, state-of-the-art, and autonomous system based on collaborative robotics. RAPTAR enables 3D radiation-pattern measurement of integrated radar modules without dedicated anechoic facilities. The system is designed to address the challenges of testing radar modules mounted in diverse real-world configurations, including vehicles, UAVs, AR/VR headsets, and biomedical devices, where traditional measurement setups are impractical. A 7-degree-of-freedom Franka cobot holds the receiver probe and performs collision-free manipulation across a hemispherical spatial domain, guided by real-time motion planning and calibration accuracy with RMS error below 0.9~mm. The system achieves an angular resolution upto $2.5^\circ$ and integrates seamlessly with RF instrumentation for near- and far-field power measurements. Experimental scans of a 60~GHz radar module show a mean absolute error of less than 2~dB compared to full-wave electromagnetic simulations ground truth. Benchmarking against baseline method demonstrates \textbf{36.5\%} lower mean absolute error, highlighting RAPTAR accuracy and repeatability.

\end{abstract}


\section{Introduction}\label{sec:I}
\IEEEPARstart{A}{ccurate} characterization of radio-frequency (RF) systems including antenna radiation patterns, total radiated power (TRP), and spectral emissions is essential for validating the performance of modern millimeter-wave (mmWave) radar and is equally applicable to other high-frequency RF sensor platforms. These systems are increasingly deployed in complex, embedded environments where they function not only as transceivers but as perception sensors, particularly in automotive radar, robotics, vehicles, and biomedical sensing \cite{richards2005fundamentals, haykin2006cognitive,bagheri2024radar}. As radar sensing becomes integral to safety-critical applications such as collision avoidance, mapping, and autonomous navigation, there is growing demand for high-resolution evaluation of their installed performance post integration and in real-world enclosures.
\begin{figure}[htbp]
\centerline{\includegraphics[width=0.49\textwidth]{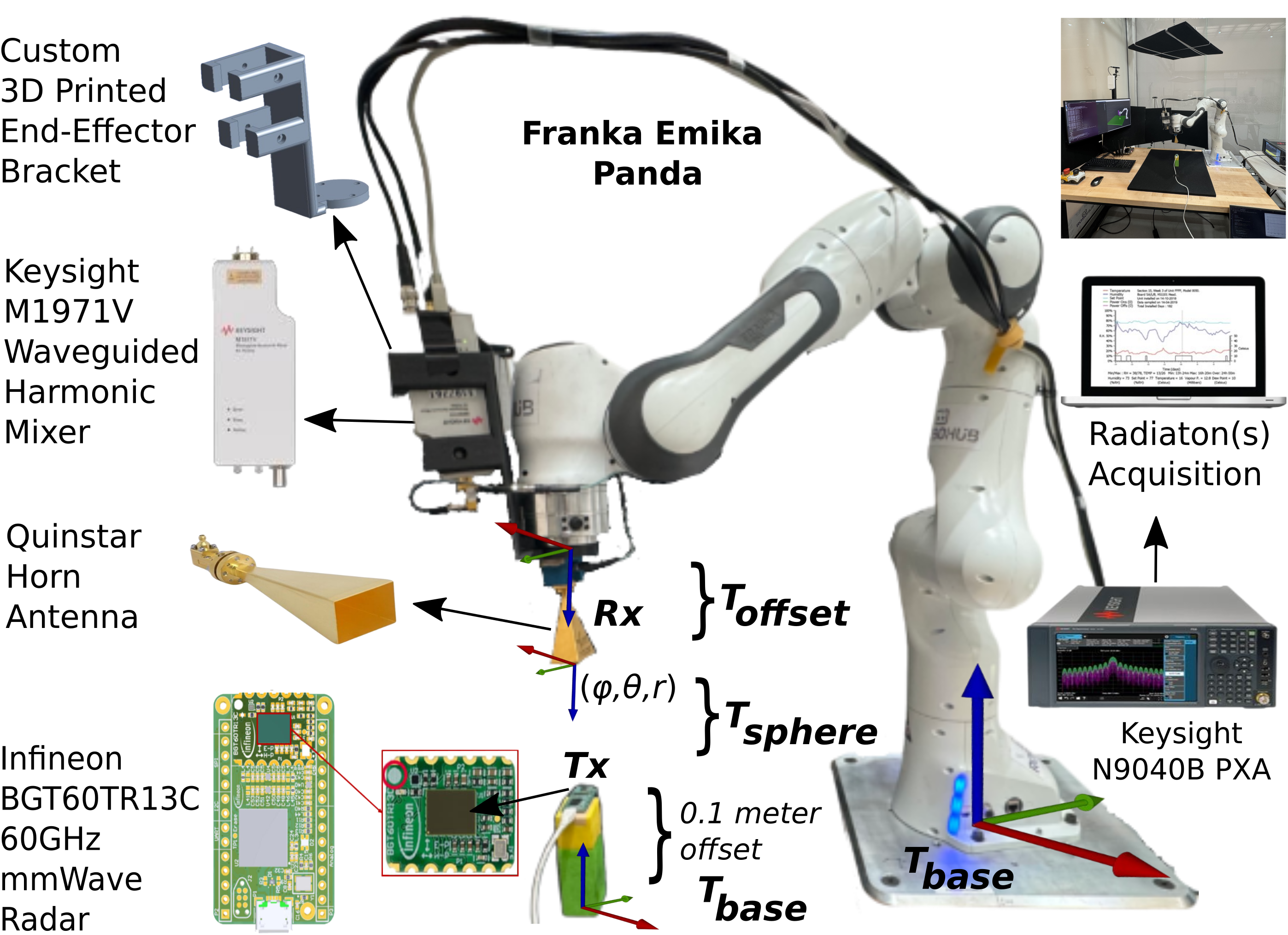}}
\caption{Applied setup overview with main attachment(s)}
\label{fig1}
\end{figure}

Traditional characterization techniques, while effective under ideal lab conditions, often fall short in capturing the real-world behavior of integrated radar systems. Enclosure effects, mutual coupling, nearby surfaces, and cable routing can significantly distort radiation patterns, shift beam lobes, or attenuate sidelobes. Existing methods typically rely on manual positioning or expensive, fixed-location motorized platforms housed in anechoic chambers \cite{balanis2015antenna, kraus2002antennas, hansen2009phased}. These setups provide high accuracy but lack flexibility, are time-intensive, and often decouple the spatial manipulation of the radar or its probe from real-time RF data acquisition, reducing throughput and increasing risk of misalignment \cite{pozar2012microwave, 8606248}.

The need for fast, reproducible, and in-situ RF testing tools is now critical not only for academic research, but also for industrial prototyping, sensor fusion validation, and regulatory compliance. In this context, collaborative robots (cobots) have emerged as an enabling technology. Lightweight and torque-sensitive, cobots such as the Franka Emika Panda can operate safely around humans and are programmable via intuitive frameworks like ROS, MoveIt!, and the Open Motion Planning Library (OMPL) \cite{bicchi2004fast, 10693652, karaman2011sampling}. Their precision, flexibility, and ease of deployment make them ideal candidates for dynamic RF measurement applications. However, their use in synchronized RF scanning especially for mmWave radar characterization remains underdeveloped.

This letter presents RAPTAR illustrated in Figure \ref{fig1}, an autonomous cobot platform designed to bridge this gap. RAPTAR enables comprehensive spatial scanning for RF measurements including radiation patterns, TRP, and emissions testing across hemispherical domains. It tightly integrates a 7-degree-of-freedom (DoF) cobot with motion planning, custom end-effector hardware, and real-time signal acquisition from spectrum analyzer(s). A key feature of RAPTAR is its suitability for evaluating \textit{installed radar performance}, where device-under-test (DUT) units are embedded in final or prototype enclosures and require high-fidelity characterization in-situ.

Unlike traditional test ranges, RAPTAR operates in semi-anechoic environments, using environmental modeling and absorbers to suppress multipath effects while ensuring cost-efficiency and deployment agility. It enables full automation—from trajectory generation to collision avoidance, pose calibration, RF triggering, and synchronized data logging.

The main contributions of this letter are:
\begin{itemize}
    \item A portable, collaborative robot-based system for mmWave RF characterization, enabling high-DoF, collision-aware scanning across 3D hemispherical domains for both near-field (NF) and far-field (FF).
    
    \item Real-time synchronization between robot poses and signal instrumentation to support automated acquisition of radiation patterns, TRP, and emissions profiles, particularly for embedded radar sensors.

    \item Environment-aware modeling and a precision 3D-printed end-effector that maintains robust horn antenna alignment across a range of radii and angular configurations.

    \item Experimental validation of the system's precision, accuracy, and reproducibility across multiple scanning modes, with quantitative comparison to both HFSS simulations and manual baselines.
\end{itemize}

The remainder of this paper is structured as follows: Section~\ref{sec:II} reviews related work. Section~\ref{sec:III} details the system architecture and calibration procedures. Section~\ref{sec:IV} presents the algorithmic and automation pipeline. Section~\ref{sec:V} discusses results. Section~\ref{sec:VI} provides insights and limitations, and Section~\ref{sec:VII} concludes with future directions.

\section{Related Work}\label{sec:II}

Parini \textit{et al.} \cite{parini2023simulation} presented a simulation-based framework for near-field measurements using a multi-axis robotic arm, leveraging phase retrieval techniques to address phase stability concerns. However, the work remained theoretical, without demonstrating an integrated system or evaluating real-world trajectory constraints. Novotny \textit{et al.} \cite{novotny2017multi} developed the LAPS, which employs two six-DoF manipulators mounted on a linear rail. While LAPS achieves dynamic over-the-air testing capabilities, the coordination of multiple robots introduces significant synchronization and calibration overhead, limiting its practicality in agile test environments.

Moser \textit{et al.} \cite{moser2024rapid} proposed a method for robotic antenna-probe alignment using iterative surface fitting and inverse kinematics, offering precision in robotic ranges. However, their setup requires extensive 3D calibration and precise external pose tracking, which increases system complexity. These solutions, although accurate, are not optimized for portability, affordability, or in-situ testing of mmWave radar sensors within real-world enclosures.

Commercial platforms such as NSI-MI Technologies' 8-axis Robotic Antenna Measurement System \cite{nsi_mi_overview, nsi2025robotic} and Boeing's DRAMS \cite{etslindgren_drams} offer robust capabilities for near-field and far-field pattern testing. The NSI-MI system accommodates antennas as large as 2.4 meters and supports a variety of test scenarios, while DRAMS enables synchronized dual-arm operation to reduce scan time. However, both platforms are expensive, infrastructure-heavy, and optimized for large-scale aerospace or defense applications. Their scalability to small-footprint labs or low-cost embedded radar modules such as those used in automotive or robotic sensing is limited.

Cobots have gained traction in manufacturing and precision assembly applications due to their safety, ease of programming, and ability to operate in shared workspaces. Keshvarparast \textit{et al.} \cite{keshvarparast2024collaborative} reviewed the integration of cobots for flexible automation, emphasizing their value in environments with payload constraints and human proximity. Sampling-based motion planning algorithms, such as those explored by Zhang \textit{et al.} \cite{zhang2025motion}, enable cobots to compute collision-free paths in constrained spaces. While these methods show promise, their direct application to synchronized mmWave RF scanning particularly in real-time environments has not been demonstrated in prior work.

In parallel, efforts to reduce dependency on full anechoic chambers have shown that semi-anechoic environments can maintain sufficient measurement fidelity when properly configured. Fernandez \textit{et al.} \cite{4463566} demonstrated that ferrite absorbers and controlled multipath environments can emulate Rayleigh fading for mobile terminal testing. Bai \textit{et al.} \cite{8471434} showed that strategic absorber placement in non-dedicated spaces can suppress multipath interference, while Breinbjerg \textit{et al.} \cite{Breinbjerg2023AntennaMC} introduced a semi-anechoic multiprobe arch for rapid automotive calibration. These works confirm that low-cost, absorber-augmented environments are viables.

Despite these advancements, there remains a clear gap in the literature: no existing system offers a low-cost, portable, and automated framework for mmWave RF testing that integrates collaborative robotics with synchronized motion planning, real-time RF data acquisition, and support for diverse modes.

\begin{figure*}[h]
\centerline{\includegraphics[width=1.0\textwidth]{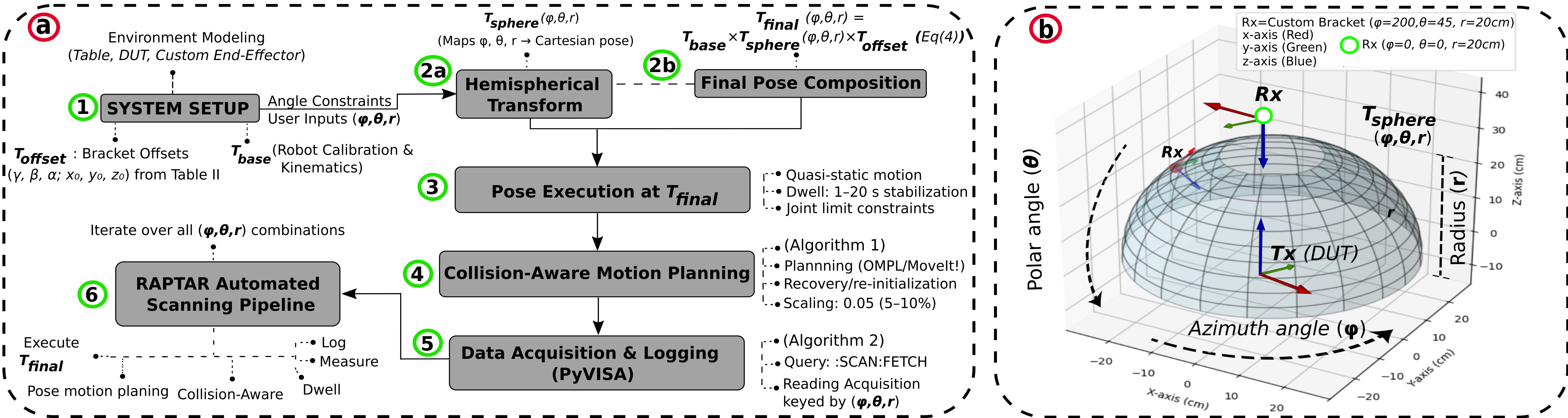}}
\caption{The left module (a) illustrates the six core components of the pipeline. On the right (b), a 3D hemispherical pattern for azimuth ($\phi$), polar ($\theta$), and radial distance ($r$). The device under test (DUT) i.e. mmWave radar (Tx), and a horn antenna (Rx)}
\label{fig2}
\end{figure*}



\section{Methods}
\label{sec:III}

This section describes the environment configuration, calibration procedures using kinematics, and the overall scanning prerequisites. We focus on how a 7-DoF cobot/manipulator (Franka Emika Panda), a custom-shaped end-effector bracket for holding the horn antenna and harmonic mixer, and collision modeling together form a robust system for conducting near-field (NF) and far-field (FF) radiation measurements. In Section \ref{sec:IV}, we address the automation and the complete scanning pipeline. Figure \ref{fig2}a shows an overall software pipeline and Figure \ref{fig2}b shows hemispherical radiation pattern.

\subsection{Environment Configuration}
\label{subsec:IIIA}

\subsubsection{Franka Emika Panda}
The Panda cobot provides a 7-DoF workspace, torque sensing, and repeatability suitable for automated scanning. We use ROS with MoveIt on Ubuntu 24.04. The robot's URDF is loaded to represent each link, including collision geometry for self-collision checks. One key constraint is the $\pm180^\circ$ limit on the final wrist joint flange\_8, which can cause local planning failures at large  angles.

\subsubsection{Table and DUT}
Our test environment features a table of dimensions $L \times W \times H$, upon which the Device Under Test (DUT) i.e. a radar is placed. We add a bounding box for this table into MoveIt! planning scene:

\begin{lstlisting}[language=Python, basicstyle=\ttfamily, breaklines=true]
planning_scene_interface.add_box("table/DUT", table_pose, (L, W, H))
\end{lstlisting}

If the DUT is placed directly on the table, polar angles ($\theta$) may be limited to avoid collisions with the bracket. Alternatively, using a riser/offset of 0.1 meters for the DUT can enable a broader range of $\theta$ (up to $\pm70^\circ$).

\subsubsection{Custom-Shaped Horn Bracket}
A custom end-effector bracket offsets the horn antenna with a yaw of $\gamma \approx -100^\circ$, minimal pitch/roll ($\beta,\alpha \approx 0$), and a few centimeters translation. This offset is encapsulated in an offset transform $\mathbf{T}_{\text{offset}}$. Manual measurement by dial gauge or 3D scanning can set these angles/translations to $\pm2$\,mm, $\pm1^\circ$ accuracy. The flange design and parameters measured from emika franka documentation and accurately designed with size to hold Rx side (Harmonic mixer) which can perform the manipulation for hemispherical pattern about Tx side (radar or DUT). 

\subsection{Robot Kinematics and Transformations}
\label{subsec:calibration}
We compute the robot’s end-effector pose relative to its base using forward kinematics, which outputs a homogeneous transformation matrix \cite{siciliano2009robotics}.

\subsubsection{Robot Base Transform}\label{subsec:Tbase}
We designate a known "stow" or "home" configuration $\{\theta_i^*\}_{i=1}^7$ with each $\theta_i^*$ denoting a specific joint angle within the robot's kinematic chain. The robot's forward kinematics at this configuration yields:

\begin{equation}\label{eq:1}
\mathbf{T}_{\text{base}} = \text{FK}(\{\theta_i^*\}), \quad \mathbf{T}_{\text{base}} \in SE(3).
\end{equation}

This base transform remains constant during a scanning session, anchoring all subsequent global manipulations. Compute the corresponding homogeneous transformation matrix from the robot's base to its flange (end-effector).

\subsubsection{Bracket Offset Calibration} \label{subsec:Toffset}
In addition to the robot's base transform, accurate characterization of the bracket's orientation and position relative to the robot flange is crucial. The custom-shaped bracket introduces an additional offset, comprising both rotational misalignments (yaw $\gamma$, pitch $\beta$, roll $\alpha$) and translational displacements $\mathbf{p}_{\text{offset}} = (x_o, y_o, z_o)$. We addressed these offsets using precision dial gauge (typical accuracies of $\pm$2 mm in translation and $\pm$1$^\circ$ in rotation) to correctly map the planned hemispherical coordinate to real-world horn antenna orientations. The resultant offset parameters are systematically encoded into a configuration file (e.g., YAML) and added in the algorithm a dedicated calibration function (such as \texttt{get\_wpose()} in the robot's software framework). Formally, the resulting homogeneous offset transform is defined as:

\begin{equation} \label{eq:2}
\mathbf{T}_{\text{offset}} = 
\begin{bmatrix}
    \mathbf{R}_{\text{offset}}(\gamma,\beta,\alpha) & \mathbf{p}_{\text{offset}}(x_o,y_o,z_o) \\
    \mathbf{0}^\top & 1
\end{bmatrix}
\end{equation}

where $\mathbf{R}_{\text{offset}}(\gamma,\beta,\alpha)$ is a rotation matrix derived from sequential Euler rotations about predefined axes. This structured and repeatable offset correction guarantees that each commanded antenna pose precisely matches its intended orientation on the spherical scanning grid.
\label{subsec:scanning}

\subsubsection{Azimuth--Polar Grid}\label{subsec:Tspere}
We define sets $\Phi = \{\phi_i\}$ i.e. $0^\circ$ to $359^\circ$, and $\Theta = \{\theta_j\}$ i.e. up to $\pm70^\circ$ max, typically with increments of minimum possible of $1^\circ$ upto max\_polar angle as shown in Figure \ref{fig2}b module.
 Each pair $(\phi,\theta)$ denotes a spherical direction from the DUT center, supplemented by a user-specified radius $r$. If the DUT is flush on the table, $\theta$ might be limited to $\pm56^\circ$ to avoid collisions. We solved this problem by putting DUT to elevated with an offset of 0.1 meters, resulted in $\Theta$ increased to $\pm70^\circ$ which provides more coverage. The complete spherical transformation with measurement coordinates coverting to cartesian space is below:

\begin{equation}\label{eq:3}
\mathbf{T}_{\text{sphere}} = 
\begin{bmatrix}
    R_z(\phi) R_y(-\theta) & r 
    \begin{bmatrix}
        \sin \theta \cos \phi \\
        \sin \theta \sin \phi \\
        \cos \theta
    \end{bmatrix} \\
    \mathbf{0}^\top & 1
\end{bmatrix}
\end{equation}



     
\subsubsection{Near-/Far-Field Radii}
We support multiple radii~$r \in \{0.03  -  0.20\}$\,m to capture both near-field and far-field coverage. NF radii ($r=0.03$\,m) can increase collision risks but provide higher resolution, while FF radii $r \leq 0.15\,\mathrm{m}$ reduce collision concerns at the cost of longer scanning times. Our pipeline seamlessly iterates over each radius while reusing the same transformation logic.

\subsubsection{Data Logging and Dwell}
For each valid pose, the system dwells 1--20\,s to stabilize mechanical vibrations. A PyVISA command i.e. \texttt{:INIT; :FETCH} retrieves power or S-parameters, storing the reading keyed by $(\phi,\theta,r)$ in a dictionary or CSV. A typical scan of 100--300 poses can last tens of minutes to hours, depending on dwell times, step sizes in polar and azimuth angles as user defines, and success rates of planning. Table \ref{table:1} shows the standardization of parameters. 

\begin{table}[htbp]
    \centering
    \caption{Standardization of Heuristics used in Methodology}
    \label{table:1}
    \resizebox{\linewidth}{!}{
        \begin{tabular}{>{\bfseries}c >{\columncolor{white}}c >{\columncolor{white}}c}
            \toprule
            \rowcolor{gray!30}\textbf{Heuristics} & \textbf{Role} & \textbf{Typical Values} \\
            \midrule
            $\gamma,\beta,\alpha$ & Bracket yaw/pitch/roll angles (degree) & $\gamma \approx -100^\circ$, $\beta = \alpha \approx 0$ \\
            $x_o,y_o,z_o$ & Translational offsets (meters) & 0.02--0.05\,m \\
            $r$ & Spherical radius & $\{0.04, 0.05, \ldots, 0.20\}$\,m \\
            $\phi$ & Azimuth angle as Phi $[1^\circ, 360^\circ]$ & $[0^\circ,1^\circ,2^\circ, \ldots, 359^\circ]$ \\
            $\theta$ & Polar angle as Theta (up to $\pm70^\circ$) & $[0^\circ,1^\circ,2^\circ, \ldots,69^\circ]$ \\
            $\nu_{\mathrm{max}}$ & Velocity scaling factor & 5--10\% of max \\
            $t_{\mathrm{dwell}}$ & Dwell time per pose & 1--20\,s \\
            \bottomrule
        \end{tabular}
    }
\end{table}

\section{Algorithm and Automation Pipeline}
\label{sec:IV}

\subsection{Final Pose Computation}
\label{subsec:VIA}

We computes the final end-effector pose $\mathbf{T}_{\text{final}}$ through a designed transformation chain that integrates measurement geometry with physical parameters. This combines three pre-requisite: base transform ($\mathbf{T}_{\text{base}}$),  spherical transform ($\mathbf{T}_{\text{sphere}}$), and offset transform ($\mathbf{T}_{\text{offset}}$) summarized in Table \ref{table:2}.

\begin{equation}\label{eq:4}
\mathbf{T}_{\text{final}} = 
\underbrace{\mathbf{T}_{\text{base}}}_{\text{Eq (\ref{eq:1})}} \times 
\underbrace{\mathbf{T}_{\text{sphere}}}_{\text{Eq (\ref{eq:2})}} \times 
\underbrace{\mathbf{T}_{\text{offset}}}_{\text{Eq (\ref{eq:3})}}
\end{equation}

Final transformation serves as the fundamental interface between the abstract measurement plan and physical robot motion, ensuring precise antenna positioning while maintaining proper boresight alignment with the DUT. Each pose calculation must account for the robot's kinematic structure, environmental constraints, and hardware imperfections to achieve the required $\pm$2 mm positioning accuracy for valid radiation pattern measurements. In the system workflow, $\mathbf{T}_{\text{final}}$ generation occurs for each measurement point $(\phi, \theta, r)$ defined in the scanning grid \ref{subsec:Tspere}. The motion planner \ref{subsec:motion_planning} utilizes these transforms to perform collision checking against the URDF model and environment, verify joint limit constraints, and generate optimized trajectories.

\subsection{Collision-Aware Motion Planning}
\label{subsec:motion_planning}

The motion planning processes each target pose $T_{\text{final}}$ through a sequential pipeline designed to ensure collision-free trajectories while maintaining measurement precision. Beginning with the computed end-effector final pose, the planner first configures MoveIt! parameters with strict velocity limits (5\% of maximum) to meet the quasi-static motion requirements . The system then performs comprehensive collision checking against both the robot's URDF model and environmental objects (table, DUT, and custom bracket) defined in Section \ref{subsec:IIIA}, utilizing OMPL's planning algorithm as the core planner. When initial planning fails typically at extreme theta angle ($\theta \geq 70^\circ$) or near joint limits the system executes a recovery protocol. This involves clearing the current target and a return to the predefined home configuration for re-planning intelligently. After successful retraction, the planner reattempts the original $T_{\text{final}}$ pose from the new configuration space region. This two-stage approach resolves edge cases while preserving the $\pm\SI{2}{\milli\meter}$ positioning accuracy required.

\begin{algorithm}[htbp]
\caption{\colorbox{gray!30}{\parbox{\dimexpr\linewidth-13\fboxsep}{Motion Planning with Recovery}}}
\SetAlgoLined
\DontPrintSemicolon
\SetKwProg{Procedure}{Procedure}{}{}
\Procedure{PlanPose {$T_{\text{final}}$, \texttt{move\_group}}}{
    \texttt{move\_group.set\_pose\_target}($T_{\text{final}}$)\;
    \texttt{move\_group.set\_max\_vel\_scaling\_fac}\;
    $\tau \gets \texttt{move\_group.plan}()$\;
    \If{$\tau.\text{success}$}{
        \Return \texttt{True} \tcp*{Primary plan success}
    }
    \Else{
        \texttt{move\_group.clear\_pose\_targets}()\;
        \texttt{move\_group.set\_named\_target}("home")\;
        $\tau_{\text{home}} \gets \texttt{move\_group.plan}()$\;
        \If {$\tau_{ \text{home}}.\text{success}$}{
            \texttt{move\_group.execute}($\tau_{\text{home}}$)\;
            \Return $\texttt{PlanPose}(T_{\text{final}}, \texttt{move\_group})$\;
        }
    }
    \Return \texttt{False}\;
}
\end{algorithm}

Key aspects of the implementation include (a) velocity Constraints fixed 5\% scaling ensures vibration control. (b) Recovery Logic Recursive retry mechanism maintains the home-position approach. (c) Validation implicit collision checking uses environmental bounds w.r.t to spatial patterns i.e. hemispherical or others. These heuristics of our approach guarantees reproducible performance across all operational scenarios while meeting the system's accuracy requirements avoiding the Joint-Limit constraints problem. Computational overhead remains minimal by leveraging MoveIt! interface.

\begin{table*}[htbp]
\caption{Transformation Components Specification}
\centering
\label{table:2}
\begin{tabular}{cccccc}
\toprule 
\rowcolor{gray!30}
\textbf{Transform} & \textbf{Role} & \textbf{Key Variables} & \textbf{Derivation / Code} & \textbf{Matrix Form} \\ \midrule
$\mathbf{T}_{\mathrm{base}}$ (Eq.~\ref{eq:1}) & 
\begin{tabular}[c]{@{}c@{}}Anchors to global coordinates\end{tabular} & 
\begin{tabular}[c]{@{}c@{}}
$\mathbf{R}_{\mathrm{base}}$: orientation matrix \\
$\mathbf{p}_{\mathrm{base}}$: position vector
\end{tabular} & 
\texttt{get\_FK()} service call & 
$\begin{bmatrix}
\mathbf{R}_{\mathrm{base}} & \mathbf{p}_{\mathrm{base}} \\
\mathbf{0}^\top & 1
\end{bmatrix}$ \\ \\

$\mathbf{T}_{\mathrm{offset}}$ (Eq.~\ref{eq:2}) & 
\begin{tabular}[c]{@{}c@{}}Encodes bracket yaw/pitch/roll\\ plus translation offsets\end{tabular} & 
\begin{tabular}[c]{@{}c@{}}
$\gamma,\beta,\alpha$: Euler angles \\
$x_o,y_o,z_o$: offset distances
\end{tabular} & 
\texttt{get\_wpose()} & 
$\begin{bmatrix}
\mathbf{R}_{\mathrm{offset}} & \mathbf{p}_{\mathrm{offset}} \\
\mathbf{0}^\top & 1
\end{bmatrix}$ \\ \\

$\mathbf{T}_{\mathrm{sphere}}$ (Eq.~\ref{eq:3}) & 
\begin{tabular}[c]{@{}c@{}} Converts spherical to Cartesian \\ coordinates (Maps $(\phi,\theta,r)$) \end{tabular} & 
\begin{tabular}[c]{@{}c@{}}
$\phi$: azimuth, 
$\theta$: polar,  
$r$: radius \\
$\mathbf{R}_{\mathrm{sphere}}$: Rotation matrix
\end{tabular} & 
\texttt{get\_A\_T\_G(phi,theta,r)} & 
$\begin{bmatrix}
\mathbf{R}_{\mathrm{sphere}} & \mathbf{p}_{\mathrm{sphere}} \\
\mathbf{0}^\top & 1
\end{bmatrix}$ \\ \\

$\mathbf{T}_{\mathrm{final}}$ (Eq.~\ref{eq:4}) & 
\begin{tabular}[c]{@{}c@{}}Executable end-effector pose\end{tabular} & 
\begin{tabular}[c]{@{}c@{}}
Composite transformation: \\
$\mathbf{T}_{\mathrm{base}}$, $\mathbf{T}_{\mathrm{sphere}}$, $\mathbf{T}_{\mathrm{offset}}$
\end{tabular} & 
\texttt{set\_pose\_target()} & 
$\mathbf{T}_{\mathrm{base}} \times \mathbf{T}_{\mathrm{sphere}} \times \mathbf{T}_{\mathrm{offset}}$ \\ \hline
\end{tabular}
\end{table*}

\subsection{Automated Measurement Acquisition Pipeline}
\label{subsec:automation}

The measurement acquisition also establishes a fully automated pipeline for RF data collection through direct instrument control which is usually done manually in the process. Upon initialization, the system configures the signal analyzer (Keysight N9040B PXA) with specified center frequency (60 GHz), span (2 GHz), and trace parameters via PyVISA's high-speed HS-LINK protocol. The core measurement cycle implements a precise sequence: first resetting the instrument to ensure consistent baseline conditions, then configuring the trace for maximum-hold detection, followed by a critical 20-second stabilization period to allow the RF signal to settle which match dwell. The system subsequently queries the analyzer for trace data, processing the raw output to extract the peak power value at the center frequency bin with 16-bit precision. Data management follows robust file handling protocols, automatically creating a new Excel workbook with appropriate headers if none exists, or seamlessly appending to an existing file while preserving previous measurements. Each acquired data point undergoes validation before being written to ensure integrity, with the system implementing atomic write operations to prevent file corruption during unexpected interruptions. The pipeline maintains strict timing control, operating at a fixed cycle time of 20 seconds per measurement (18s stabilization + 1.45s next pose plan buffer), enabling predictable total acquisition times for experiments.

\begin{algorithm}[h]
\caption{\colorbox{gray!30}{\parbox{\dimexpr\linewidth-13\fboxsep}{Automated RF Measurement Pipeline}}}
\label{alg:measurement_pipeline}
\SetAlgoLined
\DontPrintSemicolon
\textbf{Initialize:}\;
$\triangleright$ Establish TCP/IP connection via PyVISA HS-LINK\;
$\triangleright$ Configure analyzer (60GHz center, 2GHz span)\;

\BlankLine
\textbf{Measurement Loop:}\;
\For{each $i$ in $1 \rightarrow N$ measurements}{
$\triangleright$ Send :INIT:REST command\;
$\triangleright$ Set :TRAC1:TYPE MAXH\;
$\triangleright$ Wait 20 seconds (stabilization)\;
$\triangleright$ Query :FETCh:SAN1 for trace data\;
$\triangleright$ Extract center 16-bit power value\;
$\triangleright$ Validate numerical range (-100dB to +30dB)\;
$\triangleright$ Append to Excel with atomic write\;
$\triangleright$ Log timestamp, index $\theta$, $\phi$, radiation dB \;
}

\BlankLine
\textbf{Termination:}\;
$\triangleright$ Reset Tx/Rx connection with Signal Analyzer\;
$\triangleright$ Verify missing\_val; Tecplot acquisitions\;
\end{algorithm}

This pipeline eliminate(s) the impracticalities of manual data collection, particularly when characterizing thousands of antenna positions. The implementation's modular architecture enables three operational modes: standalone data acquisition, system integration via TCP triggers, and hybrid operation. A configuration file interface allows customization of measurement parameters without code modification, while the atomic write implementation prevents data loss during interruptions.

\section{Experiments}
\label{sec:V}
\begin{figure*}[htbp] 
    \centering
    \includegraphics[width=1.0\textwidth]{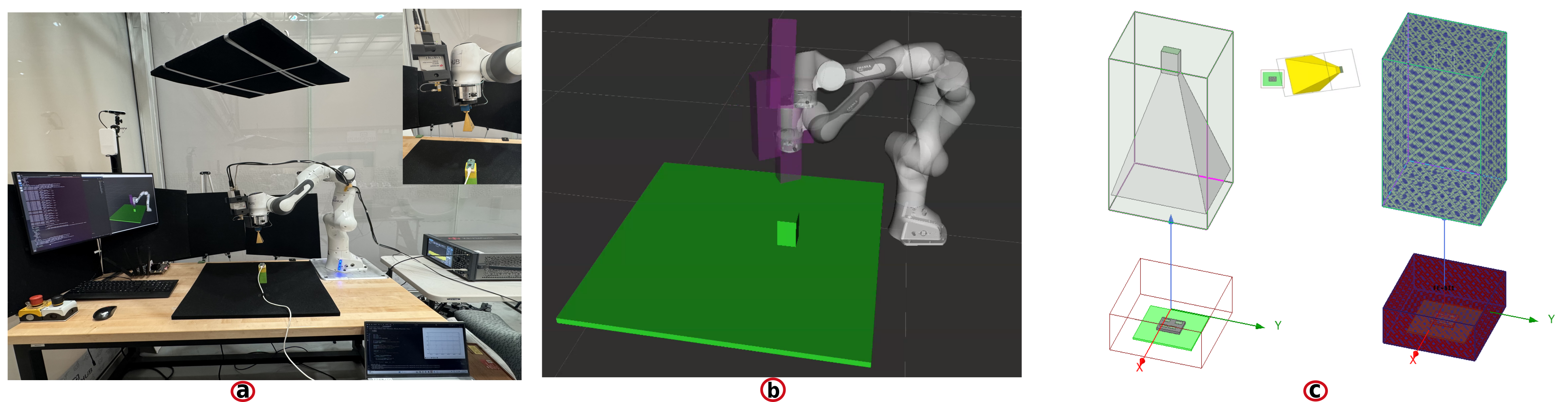}
    \caption{(a) Setup with Anechoic Absorbers (b) MoveIt! Planner RViz (c) HFSS Simulation of Scattering with FEBI Region}
    \label{fig:3}
\end{figure*}

 Exhaustive uniform sampling are selected for this study to ensure quantitative benchmarking consistency across all measurement using a Franka Emika Panda 7-DoF, Keysight N9030B PXA spectrum analyzer with M1971V harmonic mixer, BGT60TR13C mmWave radar, and a Quinstar horn antenna at 60 GHz, setup shown in Figure \ref{fig:3}. Radar is calibrated with corner reflector then over multiple days covered both H and E planes with near-field (NF) and far-field (FF) radii, testing odd radii (3, 5, 7, 9, 11, 13, 15 cm) and even radii (4, 6, 8, 10, 12, 14 cm). For even radii, measurements were taken at 15° increments; for odd radii 20° increments in both azimuth (from 0° to 360°) and polar (from 0° to 60°) angles; and an extra for 5 cm with 10° increments were sampled for max polar angle (from 0° to 70°). Each set was repeated over at least three non-consecutive days, with full shutdown before each session to assess intra- and inter-day repeatability. Motion planning are implemented using MoveIt, with the environment, DUT, and horn antenna modeled in RViz. At every $(\phi, \theta, r)$ coordinate, radiation pattern are compared with ground truth Ansys HFSS FEBI method, allows for efficient simulation of scattering problems by using a boundary integral method to truncate an open space surrounding a finite element domain and conventional manual as baseline (turntable/probe).
\vspace{-7px}
\subsection{Motion Planning and Calibration}
The RAPTAR system integrates motion planning algorithms to generate efficient, collision-free scan trajectories for spatial pattern i.e. hemi-spherical. Multiple planners including RRT-Connect, BIT*, PRM*, CHOMP, and learning-based (RL/NN) approaches were tested and benchmarked for each motion planning time, trajectory efficiency, and repeatability. Calibration accuracy and system repeatability were evaluated using two independent Panda arms (``Panda-1'' and ``Panda-2''), with intra-day and inter-day tests, and calibration drift recorded at each session shown in Table \ref{tab:3}. On-the-fly collision checks were employed during all automated scans, and no collisions or safety stops were triggered during 20 full scans.
\vspace{-7px}
\subsection{Scan Configurations and Coverage}
\begin{figure}[h]
    \centering
    \includegraphics[width=0.45\textwidth]{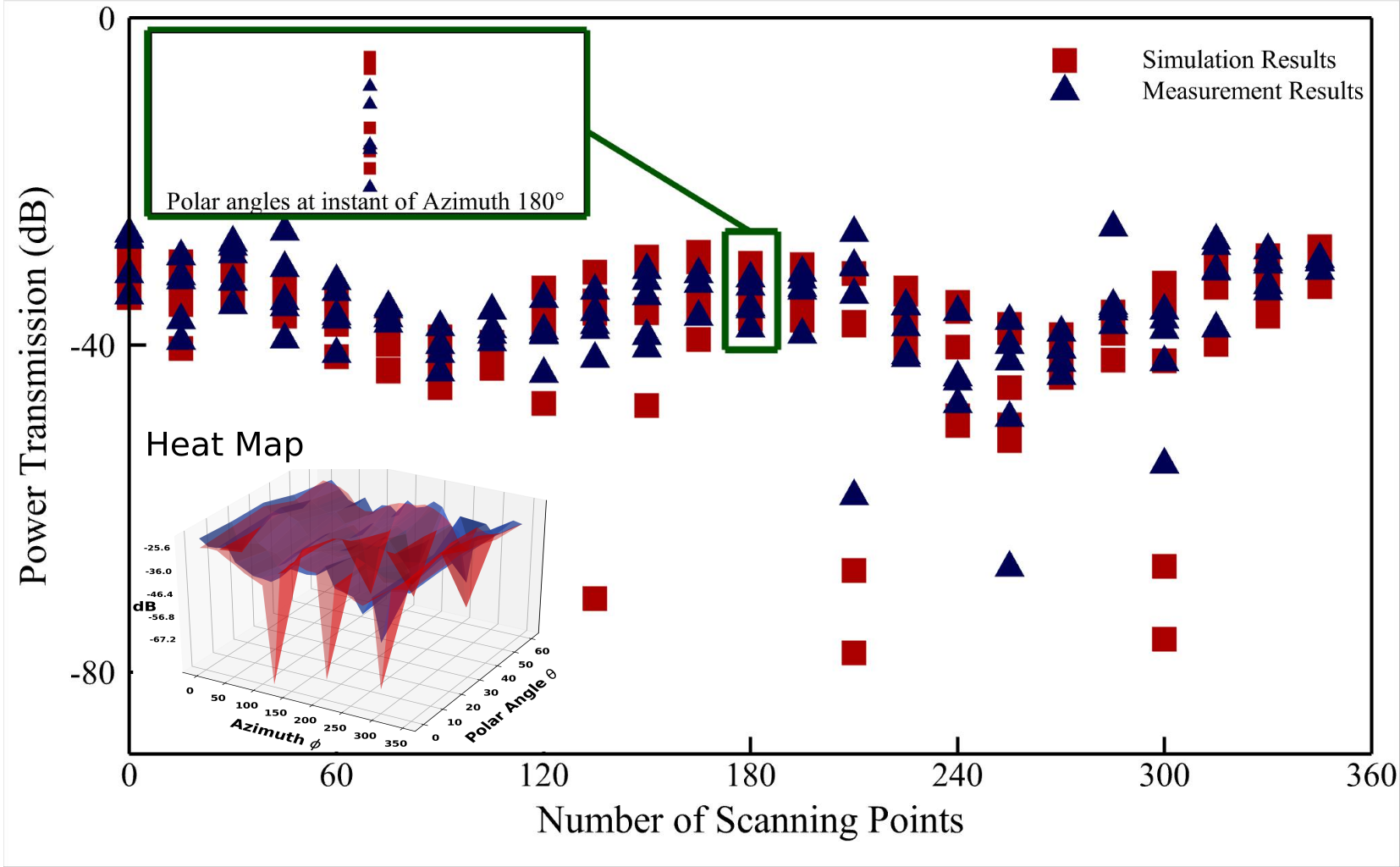}
    \caption{$r = 4\,\mathrm{cm}$, $\Delta\theta = \Delta\phi = 15^\circ$, $\theta: 0^\circ$–$60^\circ$, $\phi: 0^\circ$–$360^\circ$}
    \label{fig:4}
     \vspace*{-0.4cm}
\end{figure}

\begin{figure}[htpb]
    \centering
    \includegraphics[width=0.45\textwidth]{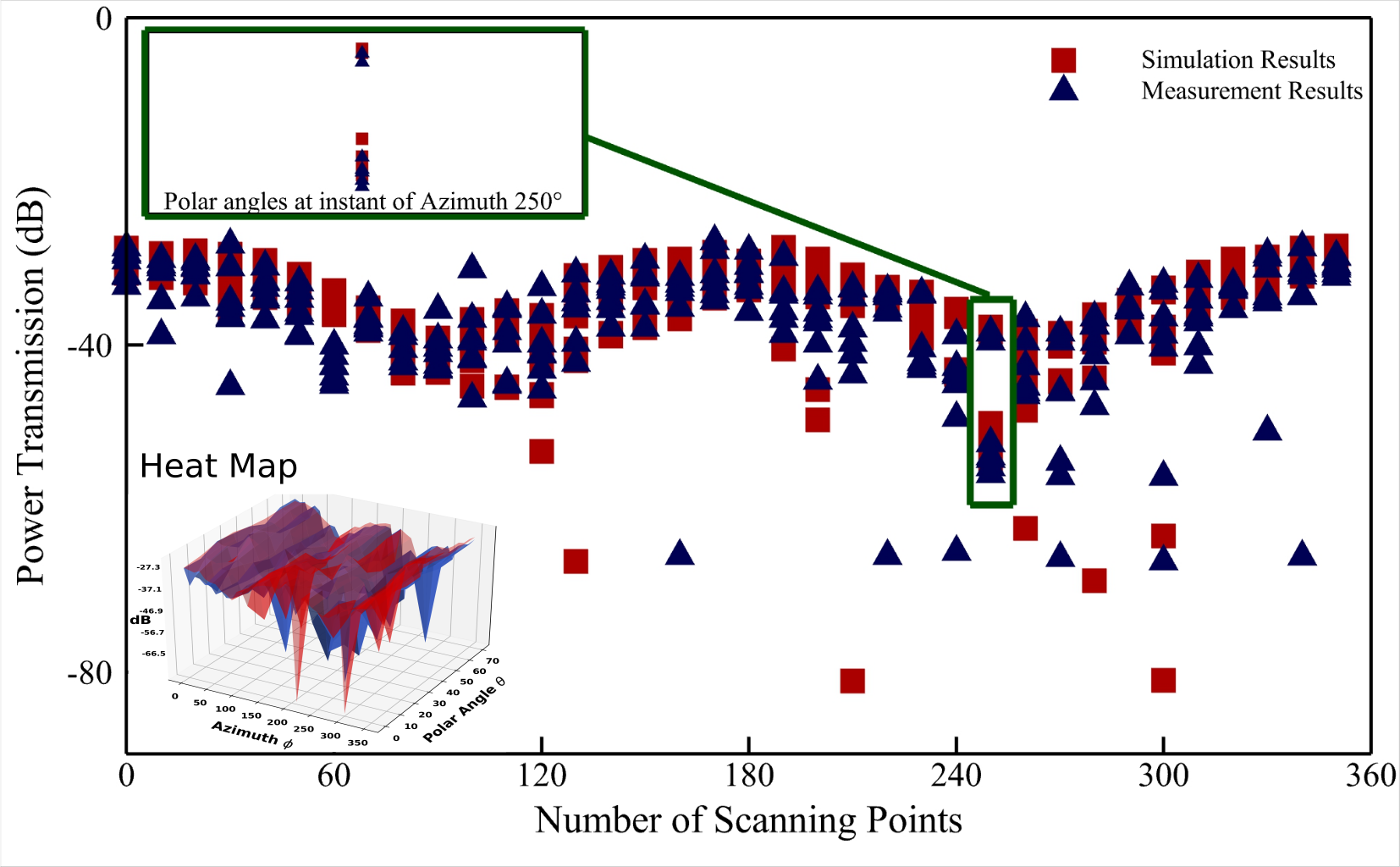}
    \caption{$r = 5\,\mathrm{cm}$, $\Delta\theta = \Delta\phi = 10^\circ$, $\theta: 0^\circ$–$70^\circ$, $\phi: 0^\circ$–$360^\circ$}
    \label{fig:5}
     \vspace*{-0.4cm}
\end{figure}

\begin{figure}[htpb]
    \centering
    \includegraphics[width=0.45\textwidth]{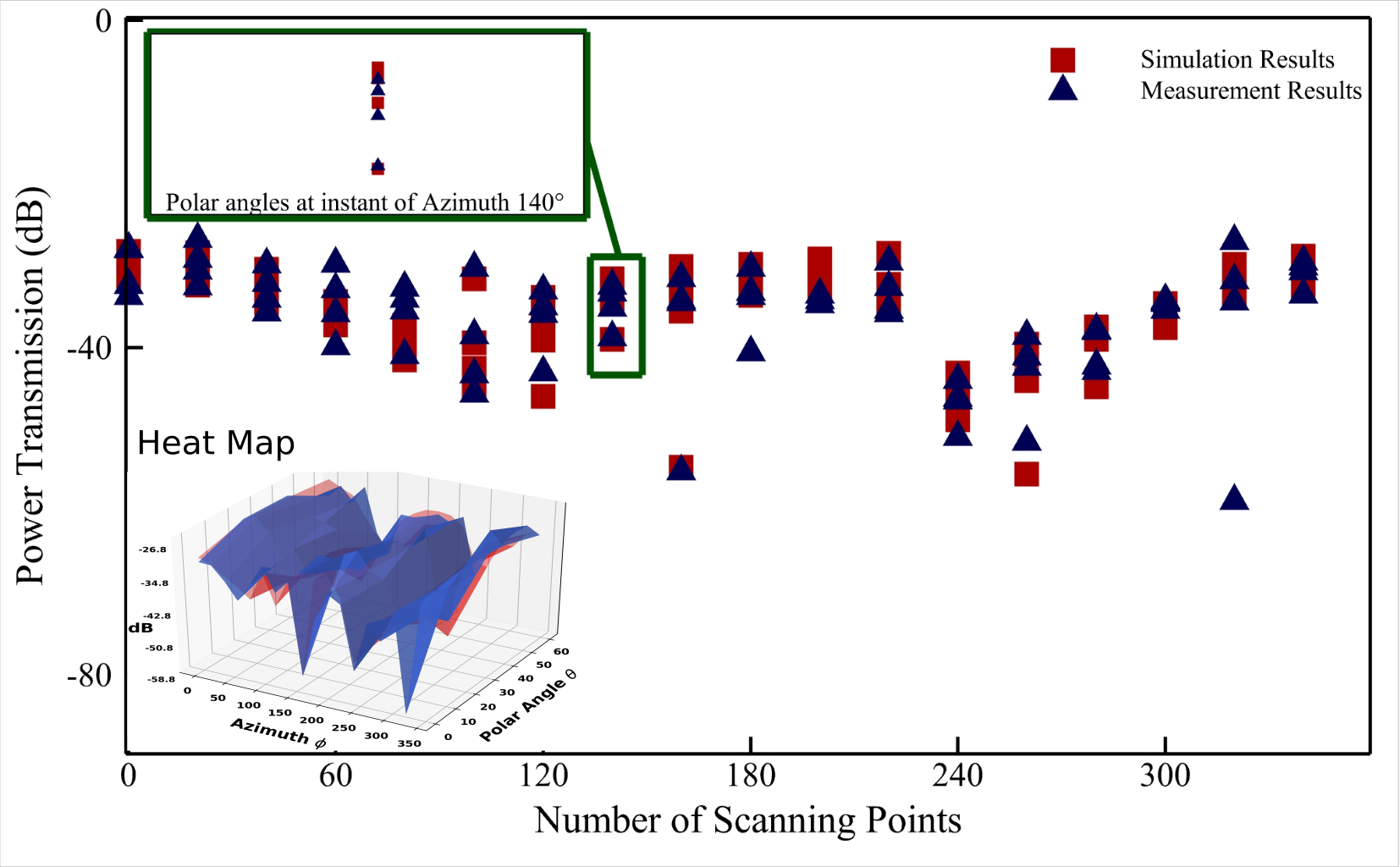}
    \caption{$r = 8\,\mathrm{cm}$, $\Delta\theta = \Delta\phi = 20^\circ$, $\theta: 0^\circ$–$60^\circ$, $\phi: 0^\circ$–$360^\circ$}
    \label{fig:6}
     \vspace*{-0.4cm}
\end{figure}

\begin{figure}[htpb]
    \centering
    \includegraphics[width=0.45\textwidth]{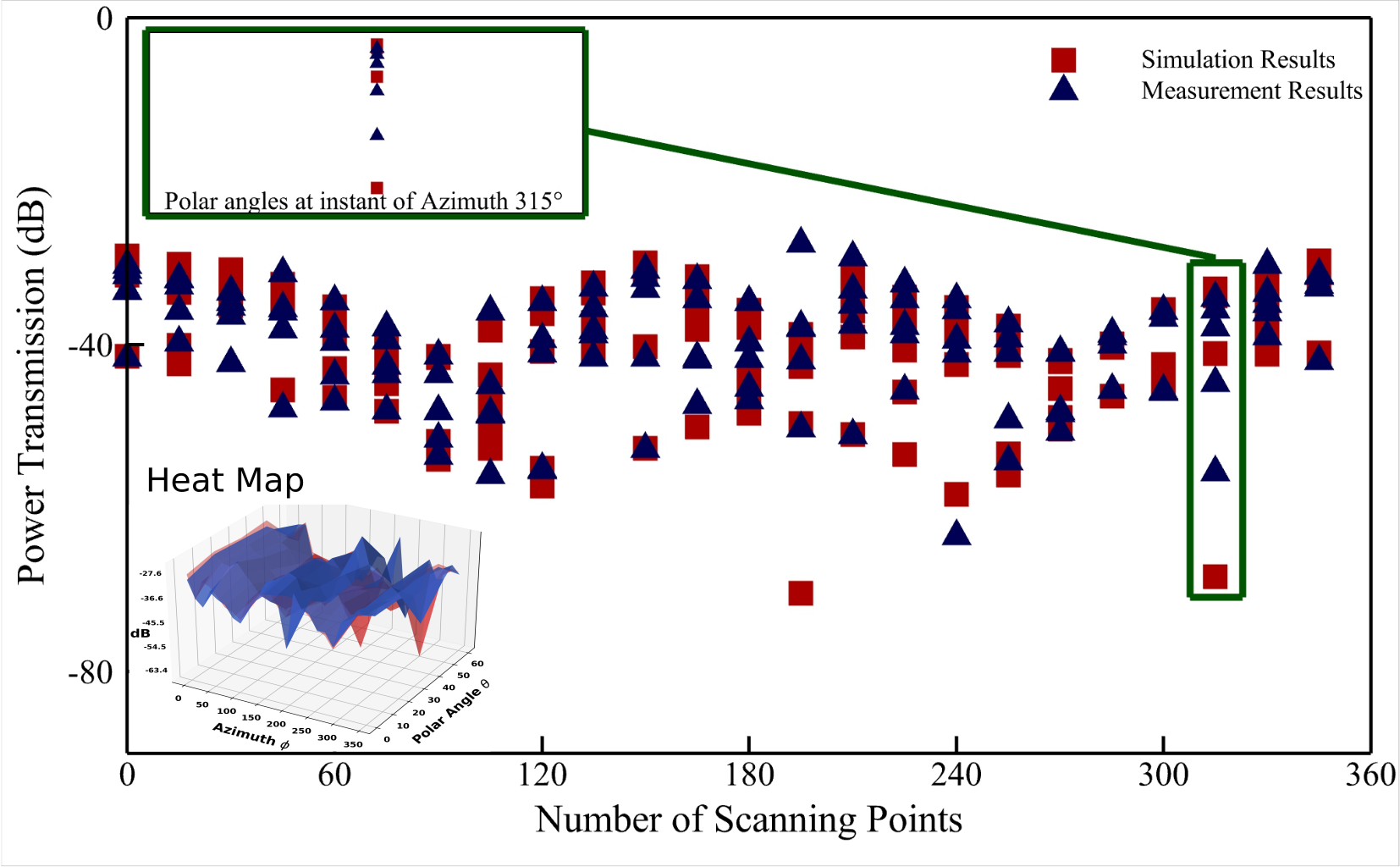}
    \caption{$r = 15\,\mathrm{cm}$, $\Delta\theta = \Delta\phi = 15^\circ$, $\theta: 0^\circ$–$60^\circ$, $\phi: 0^\circ$–$360^\circ$}
    \label{fig:7}
     \vspace*{-0.2cm}
\end{figure}

A total of thirteen scanning radii were explored in this study. For clarity and benchmarking and comparison with ground truth, four representative H-plane configurations 4\,cm (near-field), 5\,cm, 8\,cm, and 15\,cm (far-field) are summarized in Table~\ref{tab:IV} . All planned scan points were successfully acquired in each case, confirming both the collision-free motion planning and the system's reliability. Figure \ref{fig:4}, \ref{fig:5}, \ref{fig:6}, and \ref{fig:7} shows the comparison in tecplot visualization for both azimuth angle ($\phi$) with corresponding polar angles ($\theta$) at each azimuth in x-axis i.e. $\Delta\theta, \Delta\phi$ with 10° increment at each $\phi$ their are seven $\theta$ values, with 15° increment at each $\phi$ their are five $\theta$ values, and with 20° increment at each $\phi$ their are four $\theta$ values and radiation receiving at y-axis with comparison to HFSS simulation.
\begin{table*}[htbp]
\centering
\caption{Motion Planning Algorithm Benchmark and System Repeatability (with Calibration)}
\small 
\begin{tabular}{ccccccccc}
\toprule 
\rowcolor{gray!30}
 
 \textbf{Algorithm} & \textbf{Category} & \textbf{Planning} & \textbf{Success} & \textbf{Traj. Length} & \textbf{Repeatability} & \textbf{Calib. Error} & \textbf{Day-to-Day} \\
\rowcolor{gray!30}
 & & \textbf{Time} (s) & \textbf{Rate} (\%) & (m) & (MAE, dB) & (mm) & \textbf{Repeat.} (dB) \\
\midrule
RRT-Connect \cite{orthey2023sampling} & Sampling-based & 0.45 & 100 & 3.14 & 0.18 (Panda-1) & 0.37 & 0.24 \\
BIT* \cite{gammell2015batch}& Optimal Sampling & 1.5 & 100 & 2.98 & 0.21 (Panda-1) & 0.35 & 0.20 \\
PRM* \cite{orthey2023sampling}& Sampling-based & 2.8 & 100 & 3.11 & 0.22 (Panda-2) & 0.41 & 0.28 \\
CHOMP \cite{liu2022benchmarking} & Optimization & 3.2 & 98 & 2.92 & 0.15 (Panda-2) & 0.36 & 0.22 \\
PPO (Deep RL) \cite{schulman2017proximal} 
 & Learning-based & 0.62 & 100 & 3.13 & 0.19 (Panda-1) & 0.39 & 0.19 \\
\bottomrule
\end{tabular}
\label{tab:3}
\end{table*}
\begin{table}[h]
\centering
\caption{Experimental Scan Configurations and Coverage}
\begin{tabular}{cccccc}
\toprule 
\rowcolor{gray!30}
\textbf{Pattern} & \textbf{Radius} & \textbf{max\_$\theta$} & \textbf{$\theta$ Count} & \textbf{Spatial} & \textbf{Coverage} \\
\rowcolor{gray!30}
\textbf{Ratios} & (cm) & \textbf{Range} & \textbf{at Each $\phi$} & \textbf{Points} & (\%) \\
\midrule
4\,cm/$15^\circ$ & 4 & 0-60° & 5 polar & 120 & 100 \\
5\,cm/$10^\circ$ & 5 & 0-70° & 7 polar& 252 & 100 \\
8\,cm/$20^\circ$ & 8 & 0-60° & 5 polar& 72 & 100 \\
15\,cm/$15^\circ$ & 15 & 0-60° & 4 polar & 120 & 100 \\
\bottomrule
\end{tabular}\label{tab:IV}
\end{table}

\vspace{-5px}
\vspace{-5px}
\begin{table*}[h]
\centering
\caption{Quantitative Comparison Benchmark with Ground Truth and Baseline Standard Approaches}
\begin{tabular}{ccccccccc}
\rowcolor{gray!30}
\toprule 
\textbf{Pattern Ratio} & \textbf{Method} & \textbf{MAE} (\si{dB}) & \textbf{RMSE} (\si{dB}) & \textbf{Std. Error} (\si{dB}) & \textbf{SNR} (\si{dB}) & \textbf{$R^2$} & \textbf{Max Power} (\si{dB}) & \textbf{Main Lobe} (\si{\degree}) \\
\midrule
4\,cm/$15^\circ$ & RAPTAR & 1.41 & 1.71 & 1.17 & 39.5 & 0.988 & -26.3 & 15 \\
4\,cm/$15^\circ$ & Baseline & 2.22 & 2.72 & 1.31 & 33.1 & 0.945 & -26.6 & 15 \\
4\,cm/$15^\circ$ & Ground truth & --- & --- & --- & --- & 1.000 & -26.1 & 15 \\ 
\hline
5\,cm/$10^\circ$ & RAPTAR & 1.58 & 1.87 & 1.21 & 37.2 & 0.981 & -27.8 & 10 \\
5\,cm/$10^\circ$ & Baseline & 1.67 & 2.05 & 1.72 & 31.6 & 0.938 & -28.0 & 10 \\
5\,cm/$10^\circ$ & Ground truth & --- & --- & --- & --- & 1.000 & -28.1 & 10 \\ 
\hline
8\,cm/$20^\circ$ & RAPTAR & 1.23 & 1.43 & 0.98 & 41.7 & 0.995 & -29.0 & 20 \\
8\,cm/$20^\circ$ & Baseline & 1.79 & 1.94 & 1.50 & 35.2 & 0.964 & -29.4 & 20 \\
8\,cm/$20^\circ$ & Ground truth & --- & --- & --- & --- & 1.000 & -29.6 & 20 \\
\hline
15\,cm/$15^\circ$ & RAPTAR & 1.19 & 1.34 & 0.91 & 44.0 & 0.997 & -30.0 & 15 \\ 
15\,cm/$15^\circ$& Baseline & 1.52 & 2.37 & 1.20 & 36.4 & 0.971 & -30.2 & 15 \\
15\,cm/$15^\circ$ & Ground truth & --- & --- & --- & --- & 1.000 & -30.4 & 15 \\
\bottomrule
\end{tabular} 
\label{tab:V}
\end{table*}
\begin{table}[h]
\centering
\scriptsize 
\caption{Recurring Performance Metrics}
\label{tab:VI}
\begin{tabularx}{\linewidth}{X X X X X X X}
\toprule
\rowcolor{gray!30}
\textbf{Pattern Ratio} & \textbf{Repeat. (MAE, \si{dB})} & \textbf{Day-to-Day (\si{dB})} & \textbf{Success (\%)} & \textbf{Per Point (\si{s/pt})} & \textbf{Scan Time (\si{min})} & \textbf{Scan Point \si{pts/min}} \\
\midrule
4\,cm/$15^\circ$ & 0.18 & 0.24 & 100 & 20.0 & 40.0 & 3.0 \\
5\,cm/$10^\circ$ & 0.19 & 0.22 & 100 & 20.0 & 84.0 & 3.0 \\
8\,cm/$20^\circ$ & 0.17 & 0.19 & 100 & 20.0 & 24.0 & 3.0 \\
15\,cm/$15^\circ$ & 0.16 & 0.18 & 100 & 20.0 & 40.0 & 3.0 \\
\bottomrule
\end{tabularx}
\end{table}

\subsection{Quantitative Accuracy and Comparison}
RAPTAR measurements were directly compared with manual human-guided as baseline measurements and HFSS simulation as ground truth. Metrics reported in Table \ref{tab:V} include mean absolute error (MAE), root mean square error (RMSE), standard deviation, SNR, maximum gain, and main lobe direction, as well as Pearson correlation (R²). Values are averaged across three independent trials per configuration.

\begin{figure}[h]
\centerline{\includegraphics[width=0.45\textwidth]{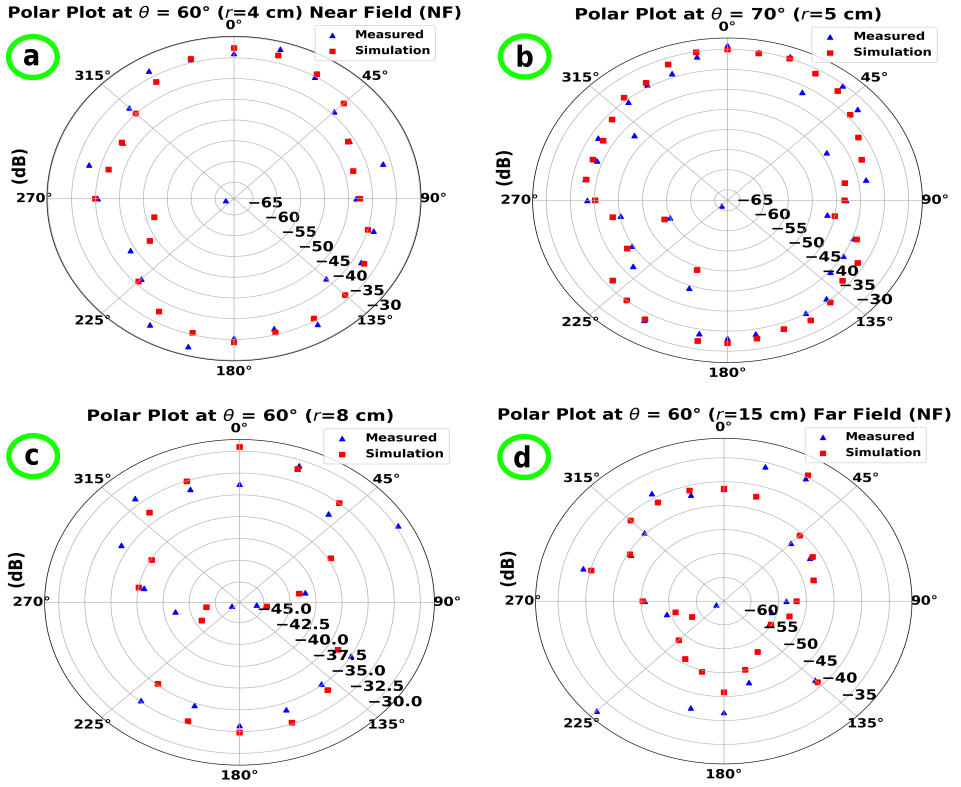}}
    \caption{ Edge Cases of $\theta$ (a) 4/$15^\circ$,(b) 5/$10^\circ$,(c) 8/$20^\circ$,(d) 15/$15^\circ$ }    \label{fig8}
    
     \vspace*{-0.2cm}
\end{figure}
\subsection{Performance Robustness, and Repeatability}
 Assessed through metrics of repeatability, day-to-day consistency, success rate, and total scan time. For each unique azimuth–polar coordinate for acquisition. This experimental protocol resulted in thousands of points, systematically covering the target hemisphere, with the entire process completed in different time w.r.t to radius. In Table~\ref{tab:VI}, no scan was aborted due to collisions and joint limit errors, demonstrating high operational reliability. Any rare missed points, primarily due to transient noise or communication dropouts, were detected and retried by the system, ensuring completeness of the dataset.
\vspace{-7 px}
\subsection{Error Distribution}
Table~\ref{tab:VII} summarizes the distribution of error, coverage, and the match between actual measured with RAPTAR  and simulated ground truths main lobe directions for the selected scans. In all configurations, 100\% of planned points were successfully acquired, and median errors were below 2 dB with near-perfect in edge-cases as well as shown in Figure \ref{fig8}.

\begin{table}[htbp]
\centering
\caption{Error Distribution across Scans}
\begin{tabular}{cccccc}
\hline
\rowcolor{gray!30} 
\textbf{Pattern Ratio} & \multicolumn{4}{c}{\textbf{Error (dB)}} & \textbf{Coverage }\\
\rowcolor{gray!30}
 & Median & Std. Dev. & Max & Min & (\%) \\
\hline
4\,cm/$15^\circ$ & 1.66 & 1.54 & 5.2 & 0.03 & 100 \\
5\,cm/$10^\circ$ & 1.85 & 1.78 & 6.1 & 0.04 & 100 \\
8\,cm/$20^\circ$ & 1.18 & 1.07 & 3.5 & 0.02 & 100 \\
15\,cm/$15^\circ$ & 1.11 & 1.02 & 2.9 & 0.01 & 100 \\
\hline
\end{tabular}
\label{tab:VII}
\end{table}

\section{Discussion and Results}\label{sec:VI}
The experimental validation of RAPTAR demonstrates not only reliable performance but also establishes a transformative benchmark for mmWave RF testing automation. Across over 20 complete 3D scan campaigns spanning varying radii, angular resolutions, and test conditions RAPTAR consistently achieved 100\% spatial coverage without a single collision, aborted scan, or system fault. This level of robustness underscores the stability of the integrated hardware-software pipeline, and affirms the system’s ability to operate reliably in extended measurement routines typical of research and industrial validation cycles.

Quantitative analysis in Table~\ref{tab:V} confirms that RAPTAR delivers sub-2\,dB mean absolute error (MAE) across all configurations, with RMSE values below 1.9\,dB and Pearson correlation coefficients ($R^2$) exceeding 0.98. Compared to manual baselines, the system not only improves accuracy but also reduces error variability and increases SNR providing consistent agreement with HFSS-based full-wave simulations. The main-lobe alignment, a critical indicator of pattern shape accuracy, remains within the angular resolution limits across all test cases. Moreover, Table~\ref{tab:VII} reveals that even in high-risk edge-angle conditions, RAPTAR maintains performance without introducing significant error outliers.

Repeatability and reproducibility are also validated: Table~\ref{tab:VI} shows intra-day and inter-day scan repeatability errors below 0.25\,dB, a threshold that rivals even high-end commercial test platforms. These results affirm RAPTAR’s suitability not only for controlled lab environments but also for field-deployable, long-duration testing.

Motion planning benchmarks in Table~\ref{tab:3} further illustrate the computational efficiency of the proposed framework. Among all evaluated algorithms, RRT-Connect emerged as the most effective for high-throughput scan planning delivering 100\% planning success rates, sub-0.5\,s computation times, and repeatability under 0.2\,dB. While optimization-based planners such as CHOMP provided marginally better trajectory efficiency, they incurred longer planning times that make them less suitable for dense, multi-radius scan campaigns.

The full-stack automation including motion control, collision recovery, synchronized RF acquisition, and data logging provides a seamless test flow with comparison to current literature. Unlike conventional platforms that require manual intervention or post-hoc data fusion, RAPTAR ensures one-to-one pose-to-measurement traceability, making it ideal for calibration, modeling, and dataset generation.

A few limitations remain. The signal analyzer’s 15–20\,s stabilization delay increases scan durations, although this can be optimized in future work via faster triggering protocols or adaptive acquisition. Boundary-angle poses (e.g., $\theta > 70^\circ$) occasionally approach joint limits and require re-planning, though the system’s automated fallback logic handles these gracefully. Finally, any changes in the probe geometry (e.g., harmonic mixer redesign) require reprinting of the end-effector bracket, although this process is modular and repeatable.


\section{Conclusions} \label{sec:VII}

This research introduced RAPTAR, an automated collaborative robot platform for mmWave radar and antenna characterization, enabling autonomous measurements in semi-anechoic environments with a 100\% scan success rate. Quantitative benchmarking shows that the system achieves up to 36.5\% lower MAE, 43.5\% lower RMSE, and 20.9\% higher SNR compared to a manual baseline approach. The $R^2$ values consistently exceed 0.98, with median errors ranging from 1.11 to 1.85~dB, and error variability (standard deviation) between 1.02 and 1.78~dB. Maximum errors remain below 6.1~dB, confirming robust and reliable performance. Among evaluated planners, RRT-Connect was identified as the optimal motion planning strategy for hemispherical scanning. RAPTAR delivers the measurement fidelity, repeatability, and automation necessary to support next-generation applications in sensing, communications, and healthcare. 

Future extensions include multi-cobot coordination for whole-body or vehicle-scale scanning and adaptive sensing through real-time feedback for future measurement infrastructure in smart factories.

\section{Acknowledgments} \label{sec:VIII}
 The authors gratefully acknowledge the support of Aidin Taeb and Keysight Technologies. Their technical guidance and equipment support were instrumental to the successful development and validation of the RAPTAR. This work was also supported by Infineon, Google, Rogers, and MITACS.


\label{sec:proposed_system}

\bibliography{reference}
\bibliographystyle{ieeetr}

\end{document}